\pdfoutput=1

\documentclass{article}
\usepackage{ijcai22}

\usepackage{times}
\usepackage{soul}
\usepackage{url}
\usepackage[hidelinks]{hyperref}
\usepackage[utf8]{inputenc}
\usepackage[small]{caption}
\usepackage{graphicx}
\usepackage{amsmath}
\usepackage{amsthm}
\usepackage{booktabs}
\usepackage{algorithm}
\usepackage{algorithmic}
\usepackage{amssymb,amsfonts}
\usepackage{multirow}
\usepackage{subfigure}
\usepackage{bm}
\usepackage{bbm}
\usepackage{balance}

\def\method{TGNN}

\title{TGNN: A Joint Semi-Supervised Framework for Graph-Level Classification}

\author{
Wei Ju$^{1,}$\thanks{Equal contribution with an alphabetical order.} \and
Xiao Luo$^{2,*}$\and
Meng Qu$^{3}$\and
Yifan Wang$^{1}$\and
Chong Chen$^{4}$\and \\
Minghua Deng$^{2,}$\thanks{Corresponding authors.}
\and 
Xian-Sheng Hua$^4$\And
Ming Zhang$^{1,\dagger}$
\affiliations
$^1$School of Computer Science, Peking University, China\\
$^2$School of Mathematical Sciences, Peking University, China\\
$^3$Mila - Québec AI Institute, Université de Montréal, Canada\\
$^4$DAMO Academy, Alibaba Group, China\\
\emails
\{juwei,xiaoluo,yifanwang,dengmh,mzhang$\_$cs\}@pku.edu.cn,
meng.qu@umontreal.ca,
\{cheung.cc,xiansheng.hxs\}@alibaba-inc.com 
}

\begin{document}

\maketitle

\begin{abstract}

This paper studies semi-supervised graph classification, a crucial task with a wide range of applications in social network analysis and bioinformatics. Recent works typically adopt graph neural networks to learn graph-level representations for classification, failing to explicitly leverage features derived from graph topology (e.g., paths). 
Moreover, when labeled data is scarce, these methods are far from satisfactory due to their insufficient topology exploration of unlabeled data.
We address the challenge by proposing a novel semi-supervised framework called Twin Graph Neural Network (TGNN). To explore graph structural information from complementary views, our TGNN has a message passing module and a graph kernel module.
To fully utilize unlabeled data, for each module, we calculate the similarity of each unlabeled graph to other labeled graphs in the memory bank and our consistency loss encourages consistency between two similarity distributions in different embedding spaces. The two twin modules collaborate with each other by exchanging instance similarity knowledge to fully explore the structure information of both labeled and unlabeled data. 
We evaluate our \method{} on various public datasets and show that it achieves strong performance.

\end{abstract}

\section{Introduction}

A fundamental problem in data mining is graph classification, which seeks to capture the property of the entire graph. The problem has been extensively studied lately with a range of downstream applications, such as predicting the quantum mechanical properties \cite{lu2019molecular,hao2020asgn} and assessing the functionality of chemical compounds \cite{kojima2020kgcn}. 
Among various graph classification algorithms~\cite{ying2018hierarchical,xu2019powerful}, message-passing neural networks (MPNNs), as the majority of graph neural networks (GNNs)~\cite{kipf2017semi,velivckovic2018graph}, have achieved tremendous success.
The main idea behind MPNNs is to propagate and aggregate messages on a graph~\cite{gilmer2017neural}, where each node receives messages from all its neighbors, and then iteratively performs neighborhood aggregation and node updates. Finally, all node representations can be combined into a graph-level representation. 
In this way, the semantic knowledge of local structures represented by different nodes can be implicitly integrated into the graph-level representation.

Although GNNs have achieved impressive performance, they do have one drawback that optimizing effective GNNs is data-hungry and frequently relies on a large number of labeled graphs, as the settings of current graph classification tasks show~\cite{xu2019powerful}.
Nevertheless, it is often expensive and time-consuming to annotate labels such as in biochemistry and chemistry~\cite{hao2020asgn}.
Although the labeled samples could be limited, massive unlabeled samples are simple to obtain, whose structures can often regularize the graph encoder to learn more discriminative representations for classification. 
This thus naturally inspires us to leverage both labeled and unlabeled data in semi-supervised scenarios to address the limitations of existing methods.


Recent efforts of combining MPNNs with semi-supervised learning techniques for graph classification \cite{li2019semi,sun2020infograph,hao2020asgn,luo2022dualgraph} are usually divided into two categories: self-training \cite{lee2013pseudo} or knowledge distillation~\cite{hinton2015distilling}. 
However, topology information is only implicitly incorporated when node features are messaging along edges, which fails to effectively explore graph topology. Moreover, as the number of labeled graphs is limited in semi-supervised graph classification, an approach that can better model graph topology and explore the unlabeled data for classification is anticipated in real-world applications.

To address the above issues, we propose a new framework named TGNN, which is capable of better exploring graph structures for semi-supervised graph-level classification. In contrast to existing MPNNs which utilize graph structures in an implicit way via message passing, our approach aims to capture graph topology in both implicit and explicit manners. On the one hand, we follow MPNNs and learn structured node representations through neighbor aggregation, and the summarized graph representation implicitly captures graph topology. On the other hand, we employ a graph kernel module, which compares each graph with parameterized hidden graphs via the random walk kernel. By this means, we get another graph representation, which leverages graph topology more explicitly. In order to couple the explicit and implicit information of graph topology, we propose a unified semi-supervised optimization framework with a novel self-supervised consistency loss, where two modules are encouraged to yield consistent similarity scores when applied to each unlabeled graph. Extensive experiments on a wide range of well-known graph classification datasets demonstrate significant improvements over competing baselines. The contributions of this work can be summarized as follows:

\begin{itemize}
	\item We provide a new semi-supervised graph classification approach called \method{}, which comprises a message passing module and a graph kernel module to explore structural information in both implicit and explicit ways.
	\item To integrate the topology information from complementary views, we propose an optimization framework including a self-supervised consistency loss to encourage exchanging similarity knowledge between two modules. 
	\item Experimental results show our \method{} significantly outperforms the compared baselines by a large margin on a variety of benchmark datasets.
\end{itemize}

\section{Related Work}
\label{sec::related}

\noindent\textbf{Graph Neural Networks.}
GNNs have been extensively investigated due to their simplification from spectral methods~\cite{spielman2007spectral} to localized models~\cite{kipf2017semi}, which connects GNNs to message passing framework~\cite{gilmer2017neural} and significantly promotes its efficiency. Most existing GNN methods for graph classification task~\cite{hao2020asgn,ju2022ghnn} inherently use message passing neural networks (MPNNs)~\cite{gilmer2017neural} to learn graph-level representation.
However, MPNNs bear the drawback that they ignore structural properties explicitly emanating from graph topology. Furthermore, recent studies have shown the inability of MPNNs to model high-order substructures such as paths~\cite{chen2020convolutional,long2021theoretically}. With our \method{}, besides implicitly learning effective graph structural information from MPNNs, we also benefit from the graph kernel module to explicitly incorporate graph topology in a semi-supervised framework.


\smallskip\noindent\textbf{Semi-supervised Graph Classification.}
Semi-supervised learning has attracted increasing attention in the field of machine learning and data mining~\cite{grandvalet2005semi,laine2017temporal,tarvainen2017mean}, whose basic idea is effectively leveraging a small number of labeled data and massive available unlabeled data to enhance the model performance. Recently much work has been done for semi-supervised graph classification~\cite{li2019semi,sun2020infograph,hao2020asgn,luo2022dualgraph}.
SEAL-AI~\cite{li2019semi} and ASGN~\cite{hao2020asgn} use active learning techniques to select the most representative samples from unlabeled data, while InfoGraph~\cite{sun2020infograph} and DualGraph~~\cite{luo2022dualgraph} fully explore the semantic information of unlabeled data via contrastive learning.
Different from existing methods which only model graph topology in an implicit way, we also consider structural information in an explicit manner via graph kernels.

\section{Methodology}

\begin{figure*}
    \centering
    \includegraphics[width=0.98\textwidth]{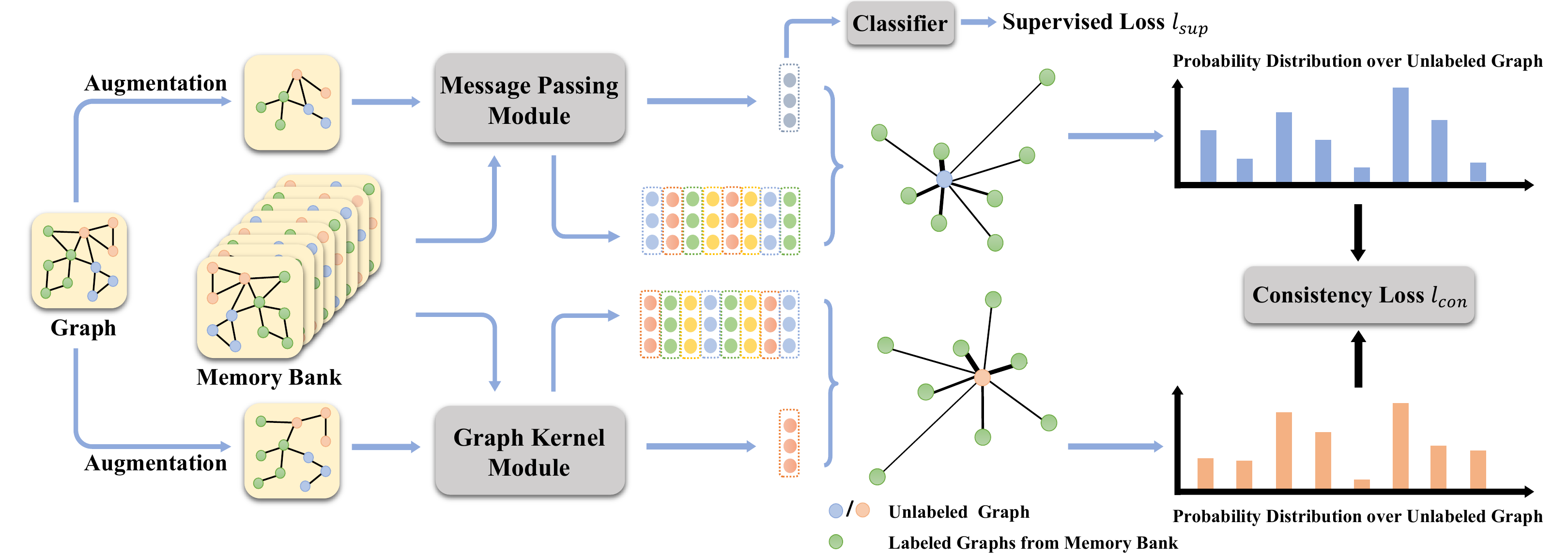}
    \caption{The schematic of the proposed framework \method{}. Our \method{} is a joint framework with twin GNNs (i.e., a message passing module and a graph kernel module), which is trained by minimizing the supervised loss as well as self-supervised consistency loss based on graph representations in two kinds of embedding spaces.}
    \label{fig::framework}
\end{figure*}

This paper provides a novel framework \method{} for semi-supervised graph classification as shown in Figure \ref{fig::framework}. At a high level, \method{} aims to capture graph topology in both implicit and explicit manners with twin GNN modules. On the one hand, we follow MPNNs to learn structured node representations via message passing, and thus the summarized graph representation reflects graph topology implicitly. On the other hand, we introduce a graph kernel module, which compares each unlabeled graph with hidden graphs via a random walk kernel to explicitly explore graph topology. To couple different topology information from twin GNNs, we present a novel semi-supervised optimization approach, in which twin modules are enforced to yield consistent similarity scores when applied to each unlabeled graph. 

\smallskip\noindent\textbf{Problem Definition.}
Let $G = (V, E)$ denote a graph, where $V$ and $E$ represent the node set and the edge set, respectively. 
In the task of semi-supervised graph classification, we are given a whole graph set $\mathcal{G} = \{\mathcal{G}^L, \mathcal{G}^U\}$, which contains labeled graphs $\mathcal{G}^L = \left\{G_{1}, \cdots, G_{\left|\mathcal{G}^{L}\right|}\right\}$ with their labels  $\mathcal{Y}^L=\{y_1, \cdots, y_{\left|\mathcal{G}^{L}\right|} \}$ and unlabeled graphs $\mathcal{G}^{U}=$ $\left\{G_{\left|\mathcal{G}^{L}\right|+1}, \cdots, G_{\left|\mathcal{G}^{L}\right|+\left|\mathcal{G}^{U}\right|}\right\}$. We aim to learn a label distribution $p(\mathcal{Y}^U |\mathcal{G}, \mathcal{Y}^L) $, which is able to predict the categories for unlabeled graphs $\mathcal{G}^U$.

\subsection{Message Passing Module}


Message passing neural networks (MPNNs) ~\cite{kipf2017semi,gilmer2017neural,velivckovic2018graph} have recently emerged as promising approaches to model graph-structured data, which implicitly encode graph structure into learned node representations $\mathbf{h}_v^{(k)}$ for each node $v$ at the $k$-th layer. MPNNs typically use neighborhood aggregation to iteratively update the representation of a node by incorporating representations of its neighboring nodes in the previous layer. After $k$ iterations, a node representation $\mathbf{h}_v^{(k)}$ is able to capture the structural information within its $k$-hop neighborhoods. Formally, the $k$-th layer of a node representation $\mathbf{h}_v^{(k)}$ obtained from the MPNN is formulated as:

\begin{equation}
\mathbf{h}_{v}^{(k)}= \operatorname{COM}^{(k)}_{\theta}\left(\mathbf{h}_{v}^{(k-1)} + \sum_{u \in \mathcal{N}(v)}\mathbf{h}_{u}^{(k-1)} \right)
\end{equation}
Here $\mathcal{N}(v)$ represents the set neighbors of node $v$, $\operatorname{COM}^{(k)}_{\theta}$ is a non-linear combination operations at layer $k$. Eventually, by using an attention mechanism, the graph-level representation will be generated by integrating all node representations at layer $K$ as follows:

\begin{equation}
f_{\theta}\left(G\right)=\sum_{v\in V} a_v \mathbf{h}_v^{(K)}
\end{equation}
\begin{equation}
a_v = \frac{\text{exp}(\mathbf{t}^{\top} \mathbf{h}_v^{(K)}) \mathbb{I}_{\mathbf{t}^{\top} \mathbf{h}_v^{(K)}>0}}{\sum_{v\in V}\text{exp}(\mathbf{t}^{\top} \mathbf{h}_v^{(K)}) \mathbb{I}_{\mathbf{t}^{\top} \mathbf{h}_v^{(K)}>0}}
\end{equation}
where $\mathbf{t}$ is a learnable vector, $f_{\theta}\left(G\right)$ is the graph-level representation, and $\theta$ denotes the parameters of the message passing module. $\mathbb{I}_{(\cdot)}$ is an indicator equaling 1 when the condition is met, and its introduction is to adaptively prune some unimportant nodes for discriminative graph-level representations.

\subsection{Graph Kernel Module}
 

However, recent MPNNs simply capture structural information implicitly via message-passing along edges. It is usually difficult to explore rich high-order substructures (e.g., paths~\cite{chen2020convolutional,long2021theoretically}) in various MPNN models. It is therefore desirable to have a principled way of incorporating structural information explicitly.

To this end, we propose to use a random walk graph kernel to explicitly explore graph topology. Our graph kernel module introduces some hidden graphs which are parameterized using trainable adjacency matrices. Formally, our module contains $N$ hidden graphs $G'_1, G'_2, \cdots, G'_N$ with different sizes. We parametrize each hidden graph $G'_i$ as an undirected graph for fewer parameters. These hidden graphs are expected to learn structures that help distinguish between available classes. 
Inspired by the fact that the random walk kernels quantify the similarity of two graphs based on the number of common walks in the two graphs~\cite{kashima2003marginalized,nikolentzos2020random},
we compare each graph sample against hidden graphs with a differentiable function from the random walk kernel. 

Specifically, given two graphs $G=(V,E)$ and $G'=(V',E')$, their graph direct product $G_\times = (V_\times, E_\times) $ is a graph where \(V_{\times}=\left\{\left(v, v^{\prime}\right): v \in V \wedge v^{\prime} \in V^{\prime}\right\}\) and \(E_{\times}=\left\{\left\{\left(v, v^{\prime}\right),\left(u, u^{\prime}\right)\right\}:\{v, u\} \in E \wedge\left\{v^{\prime}, u^{\prime}\right\} \in E^{\prime}\right\}\). It could be observed that a random walk on $G_\times$ can be interpreted as a simultaneous walk on graphs $G$ and $G'$~\cite{vishwanathan2010graph}. Considering traditional random walk kernels count all pairs of matching walks on $G$ and $G'$, the number of matching walks is equivalent to the adjacency matrix $\mathbf{A}_\times$ of $G_\times$ if traversing over nodes of $G$ and $G'$ at random. The $P$-step ($P \in \mathbb{N}$) random walk kernel between $G$ and $G'$ that counts all simultaneous random walks is thus defined as:

\begin{equation}
	k\left(G, G^{\prime}\right)=\sum_{p=0}^{P} \omega_{p} \mathbf{e}^{\top} \mathbf{A}_{\times}^{p}\mathbf{e}
\end{equation}
where $\mathbf{e}$ is an all-one vector, and $\omega_0, \dots, \omega_P$ are positive weights. To simplify the calculation, we only compute the number of common walks of length exactly $p$ over two compared graphs:
\begin{equation}
k^{(p)}\left(G, G^{\prime}\right)= \mathbf{e}^{\top} \mathbf{A}_{\times}^{p}\mathbf{e}
\end{equation}
Then, given $\mathcal{P} = \{0,\dots, P \}$ and hidden graph set $\mathbf{\mathcal{G}}_h = \{G'_1, \dots, G'_N\}$, we can build a matrix $\mathbf{H}\in \mathbb{R}^{N\times (P+1)}$ where $\mathbf{H}_{ij} = k^{(j-1)}(G, G'_i)$ for each input graph $G$. Finally, the matrix $\mathbf{H}$ is flattened and fed into a fully-connected layer to produce the graph-level representation denoted as $g_\phi(G)$.


\subsection{Semi-supervised Optimization Framework}
In this subsection, we discuss how to integrate the twin modules to explore graph structural information from complementary views in semi-supervised scenarios. 

To produce generalized and robust representations for two modules, we first involve four fundamental data augmentation strategies to create positive views of graphs that preserve intrinsic structural and attribute information: (1) \emph{Edge dropping} (2) \emph{Node dropping} (3) \emph{Attribute masking}  (4) \emph{Subgraph}. For more details of these strategies, refer to \cite{you2020graph}.
We generate augmented graphs by randomly selecting one of the defined augmentation strategies. Specifically, given an instance $G_j$, for two modules, two stochastic augmentations $T^1$ and $T^2$ sampled from their corresponding augmentation families are applied to $G_j$, resulting in two correlated samples denoted as $G_j^1 = T^1(G_j)$ and $G^2_j = T^2(G_j)$. 
Afterward, $G_j^1$ and $G_j^2$ are fed into two modules respectively to obtain the embedding pair $(z_j=f_\theta(G_j^1), w_j = g_\phi(G_j^2))$.

Since two modules mine semantic information from GNNs and graph kernels respectively, the semantic representations derived from their respective representation spaces might exist a discrepancy. Moreover, since the label annotations are usually limited in a semi-supervised case, predicted pseudo labels are usually unreliable and biased.
As such, simply aligning the graph-level representations or pseudo labels of the same instance may be sub-optimal, especially for unlabeled samples. To alleviate the issue, we propose to enhance each instance by exchanging instance knowledge via comparing its similarities to other labeled samples in embedding spaces of two modules.

Specifically, we first randomly select a set of the labeled data $\{G_{a_1},\dots,G_{a_M}\}$ as anchor samples that are stored in a memory bank and then embed them using both message passing module and graph kernel module to get their representations $\{z_{a_m}\}_{m=1}^M$ and $\{w_{a_m}\}_{m=1}^M$. 
Note that we need a large set of anchor samples so that they have large variations to cover the neighborhood of any unlabeled sample. However, it is computationally expensive to process many samples in a single iteration due to limited computation and memory. To tackle this issue, we maintain a memory bank as a queue defined on the fly by a set of anchor samples from several most recent iterations for our twin GNN modules. 
 
In detail, given an unlabeled graph $G_j$, we get its embeddings $z_j=f_\theta(T^1(G_j))$ via the message passing module, and then we calculate the pairwise similarity between $z_j$ and all anchor embeddings $\{z_{a_m}\}_{m=1}^M$. Similarly, the pairwise similarity for the graph kernel module can be obtained through comparison between $w_j=g_\phi(T^2(G_j))$ and $\{w_{a_m}\}_{m=1}^M$. In our implementation, we use exponential temperature-scaled cosine similarity to measure the relationship in both embedding spaces. Formally, the similarity distribution between the unlabeled sample and anchor samples in the message passing module's embedding space is: 
\begin{equation}
p_{m}^{j}=\frac{\exp \left( \cos(z_j, z_{a_m}) / \tau\right)}{\sum_{m'=1}^{M} \exp \left( \cos (z_j, z_{a_{m'}}) / \tau\right)}
\end{equation}
where $\tau$ is the temperature parameter set to $0.5$ following \cite{you2020graph} and $\cos(a,b)=
\frac{a \cdot b}{\left\|a\right\|_{2}\left\|b\right\|_{2}}
$ is the cosine similarity. In the same way, the similarity in the embedding space of the graph kernel module is written as:
 \begin{equation}
q_{m}^{j}=\frac{\exp \left( \cos(w_j, w_{a_m}) / \tau\right)}{\sum_{m'=1}^{M} \exp \left( \cos (w_j, w_{a_{m'}}) / \tau\right)}
\end{equation}

For each unlabeled graph $G_j$, we encourage the consistency between probability distributions $p^{j}=[p_{1}^{j}, \dots, p_{M}^{j}]$ and $q^{j}=[q_{1}^{j}, \dots, q_{M}^{j}]$ by using Kullback-Leibler (KL) Divergence as the measure of disagreement. In formulation, we present a consistency loss defined as follows:
\begin{equation}
\mathcal{L}_{con}=\frac{1}{|\mathcal{G}^U|}\sum_{G_j \in \mathcal{G}^U } \frac{1}{2}(D_{\mathrm{KL}}(p^j \| q^j)+ D_{\mathrm{KL}}(q^j \| p^j))
\end{equation}

\begin{algorithm}[t]
    \caption{\method{}’s main learning algorithm}
    \label{alg}
    \textbf{Input}: Labeled data $\mathcal{G}^L$, unlabeled data $\mathcal{G}^U$ \\
    \textbf{Parameter}: Message passing module parameter $\theta$, graph kernel module parameter ${\phi}$, classifier parameter $\eta$  \\
    \textbf{Output}: classifier $\Phi\left(y|G\right)$ 
    \begin{algorithmic}[1] 
        \STATE Sample $M$ examples from $\mathcal{G}^L$ to construct anchor set as the memory bank.
        \STATE Initialize parameters $\{\theta,\phi,\eta\}$.
        \WHILE{not convergence}
            \STATE Sample minibatch $\mathcal{B}^L$ and $\mathcal{B}^U$.
            \STATE Forward propagation $\mathcal{B}^L$ and $\mathcal{B}^U$ via twin modules.
            \STATE Compute objective function using Eq. \eqref{eq:loss}.
            \STATE Update the parameters by back propagation.
            \STATE Update the memory bank with $\mathcal{B}^L$ for two modules following a first-in, first-out manner.
        \ENDWHILE 
    \end{algorithmic}
\end{algorithm}

In order to output label distribution for classification, we use the graph-level representation from the message passing module to predict the label through a multi-layer perception (MLP) classifier $\mathcal{H}_\eta(\cdot)$ (i.e., $\Phi(y \mid G)=\mathcal{H}_{\eta}(f_\theta(G) )$. We choose the message passing module since a single MPNN outperforms a single graph kernel network empirically by validation study. Therefore, we use the cross-entropy function to characterize the supervised classification loss: 
\begin{equation}
\mathcal{L}_{sup}=\frac{1}{|\mathcal{G}^L|}\sum_{G_j \in \mathcal{G}^L }\left[-\log \Phi(y_j \mid G_j)\right]
\end{equation}
Finally, we combine the supervised classification loss $\mathcal{L}_{sup}$ with unsupervised consistency loss $\mathcal{L}_{con}$ in the overall loss:
\begin{equation}\label{eq:loss}
\mathcal{L}=\mathcal{L}_{sup}+\mathcal{L}_{con}
\end{equation}
The overall framework is illustrated in Algorithm \ref{alg}. 

\begin{table*}[ht]
\centering
\tabcolsep=6.5pt
\begin{tabular}{lccccccc}
\toprule
{Methods} &  {PROTEINS} & {DD}   & {IMDB-B} & {IMDB-M}  & {REDDIT-B} & {REDDIT-M-5k} & {COLLAB} \\
\midrule 
WL  & $ 63.5\pm1.6 $ & $ 57.3\pm1.2 $  & $ 58.1\pm2.3  $ & $  33.3\pm1.4  $  & $  61.8\pm1.3  $  & $ 37.0\pm0.9  $ & $ 62.9\pm0.7  $ \\
Sub2Vec & $ 52.7\pm4.5 $ & $ 46.4\pm3.2 $   & $ 44.9\pm3.5  $ & $  31.8\pm2.7  $  & $  63.5\pm2.3  $  & $ 35.1\pm1.5  $ & $ 60.8\pm1.4  $ \\
Graph2Vec  & $ 63.1\pm1.8 $ & $ 53.7\pm1.6 $  & $ 61.2\pm2.6  $ & $  38.1\pm2.2  $  & $  67.7\pm2.3  $  & $ 38.1\pm1.4  $ & $ 63.6\pm0.9  $ \\
\midrule
EntMin & $ 62.7\pm2.7 $ & $ 59.8\pm1.3 $    & $ 67.1\pm3.7 $ & $ 37.4\pm1.2 $  & $ 66.9\pm3.5 $ & $ 38.7\pm2.8 $ & $ 63.8\pm1.6 $  \\
Mean-Teacher  & $ 64.3\pm2.1 $ & $ 60.6\pm1.8 $   & $ 66.4\pm2.7 $ & $ 38.8\pm3.6 $  & $ 68.7\pm1.3 $ & $ 39.2\pm2.1 $  & $ 63.6\pm1.4 $\\
VAT  & $ 64.1\pm1.2 $ & $ 59.9\pm2.6 $   & $ 67.2\pm2.9 $ & $ 39.6\pm1.4 $  & $ 70.8\pm4.1 $  & $ 38.9\pm3.2 $ & $ 64.1\pm1.1 $ \\
\midrule
InfoGraph  & $ 68.2\pm0.7 $& $ 67.5\pm1.4 $  & $ 71.8\pm2.3 $ &  $ 42.3\pm1.8 $  & $ 75.2\pm2.4 $ & $ 41.5\pm1.7 $ & $ 65.7\pm0.4 $ \\
ASGN  & $ 67.7\pm1.2 $ & $ 68.5\pm0.6 $  & $ 70.6\pm1.4 $ & $ 41.2\pm1.4 $  & $ 73.1\pm2.3 $ & $ 42.2\pm0.8 $ & $ 65.3\pm0.8 $\\
GraphCL & $ 69.4\pm0.8 $ & $ 68.7\pm1.2 $ & $ 71.2\pm2.5 $ & $ 43.7\pm1.3 $ & $ 75.2\pm1.7 $ & $ 42.3\pm0.9 $ & $ 66.4\pm0.6 $\\
JOAO & $ 68.7\pm0.9 $ & $ 67.9\pm1.3 $ &  $ 71.0\pm1.9 $ & $ 42.6\pm1.5 $ & $ 74.8\pm1.6 $ & $ 42.1\pm1.2 $ & $ 65.8\pm0.4 $\\
DualGraph  &  $ 70.1\pm1.2 $  & $ 69.8\pm0.8 $  &  $ 72.1\pm0.7 $  &  \textbf{44.8 $\pm$ 0.4} &  $ 75.4\pm1.4 $  &  $ 42.9\pm1.4 $  &  $ 67.2\pm0.6 $ \\
\midrule 
\method{} (Ours)  &  \textbf{71.0 $\pm$ 0.7}  &  \textbf{70.8 $\pm$ 0.9}   &  \textbf{72.8 $\pm$ 1.7}  &  42.9 $\pm$ 0.8     &  \textbf{76.3 $\pm$ 1.3}  &  \textbf{43.8 $\pm$ 1.0}  &  \textbf{67.7 $\pm$ 0.4} \\
\bottomrule
\end{tabular}
\caption{Quantitative results of different algorithms. We highlight that our \method{} outperforms all other baselines on most datasets.}
\label{tab::results}
\end{table*}

\section{Experiments}
\label{sec::experiment}

\subsection{Experimental Setups}

\smallskip\noindent\textbf{Benchmark Datasets.}
We evaluate our proposed \method{ } using seven publicly accessible datasets (i.e., PROTEINS, DD, IMDB-B, IMDB-M, REDDIT-B, REDDIT-M-5k and COLLAB~\cite{yanardag2015deep}) and two large-scale OGB datasets (i.e., OGB-HIV, OGB-MUV). 
Following DualGraph~\cite{luo2022dualgraph}, we adopt the same data split, in which the ratio of labeled training set, unlabeled training set, validation set and test set is 2:5:1:2.



\smallskip\noindent\textbf{Competing Models.}
We carry out comprehensive comparisons with methods from three categories: classical graph approaches (i.e., WL~\cite{shervashidze2011weisfeiler}, Sub2Vec~\cite{adhikari2018sub2vec} and Graph2Vec~\cite{narayanan2017graph2vec}), classical semi-supervised learning approaches (i.e., EntMin~\cite{grandvalet2005semi}, Mean-Teacher~\cite{tarvainen2017mean} and VAT~\cite{miyato2018virtual}) and graph-specific semi-supervised learning approaches (i.e., InfoGraph~\cite{sun2020infograph}, ASGN~\cite{hao2020asgn}, GraphCL~\cite{you2020graph}, JOAO~\cite{you2021graph}), and DualGraph~\cite{luo2022dualgraph}. 


\smallskip\noindent\textbf{Implementation Details.}
All methods are implemented in PyTorch, and the experiments are presented with the average prediction accuracy (in $\%$) and standard deviation of five times. For the proposed \method{}, we empirically set the embedding dimension to $64$, the number of epochs to $300$, and batch size to $64$. We modify GIN~\cite{xu2019powerful} to parameterize the message passing module $f_\theta$, consisting of three convolution layers and one pooling layer with an attention mechanism. 
For our graph kernel module $g_\phi$, we empirically set the number of hidden graphs to $16$ and their size equal to $5$ nodes. The maximum length of random walk $P$ is set to $3$. 
Finally, we use Adam to optimize all the models.

\subsection{Results and Analysis}
Table~\ref{tab::results} shows the results of semi-supervised graph classification using half of the labeled data. We can get the following observations: 
(1) The majority of the classical graph approaches achieve worse performance than others, indicating that GNNs have a high representation capability to extract effective information from graph data. 
(2) The approaches with classical semi-supervised learning techniques show worse performance compared with the graph-specific semi-supervised learning approaches, demonstrating that models specifically designed for graphs can better learn the characteristics of graphs in semi-supervised scenarios. 
(3) Our framework \method{} outperforms all other baselines on six of seven datasets, soundly justifying the superiority of our framework. 
Moreover, we assess our model on large-scale OGB datasets and the results on Table~\ref{tab::results_ogb} further validate the effectiveness of \method{}.

\begin{table}[t]
\centering
\tabcolsep=7pt
\begin{tabular}{@{}llcc@{}}
\toprule
{Rate}   & {Methods} & {OGB-HIV} & {OGB-MUV}  \\
\midrule 
\multirow{3}*{1\%} & InfoGraph  & $ 51.3\pm3.8 $ & $ 50.1\pm1.6 $ \\
& GraphCL  & $ 45.1\pm5.7 $ & $ 50.8\pm1.3 $  \\
\cmidrule{2-4}
& \method{} (Ours) &  \textbf{53.3 $\pm$ 4.1}  &   \textbf{52.8 $\pm$ 1.5}  \\
\midrule
\multirow{3}*{10\%} & InfoGraph  & $ 63.7\pm1.2 $ & $ 51.3\pm0.8 $ \\
& GraphCL  & $ 63.2\pm0.5 $ & $ 51.1\pm0.7 $  \\
\cmidrule{2-4}
& \method{} (Ours) &  \textbf{64.1 $\pm$ 0.8}  &   \textbf{53.5 $\pm$ 0.7}  \\
\bottomrule
\end{tabular}
\caption{Results on large-scale OGB datasets. (Test ROC-AUC on OGB-HIV, OGB-MUV at $1\%$ and $10\%$ label rate respectively.)}
\label{tab::results_ogb}
\end{table}

\begin{figure}[t]\small
\centering
\subfigure[PROTEINS]{
\centering
\includegraphics[width=0.225\textwidth]{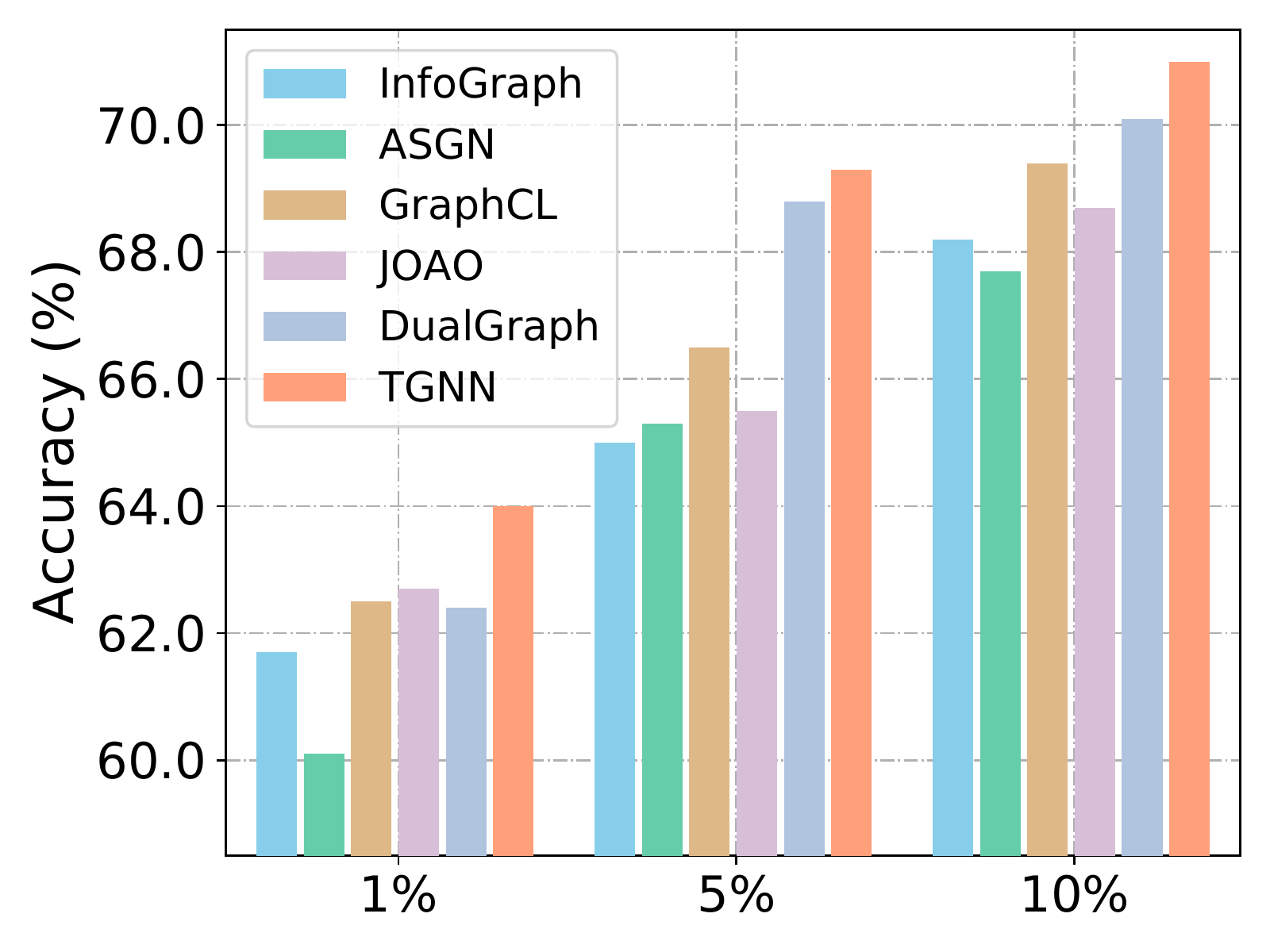}
}
\hspace{-2mm}
\subfigure[DD]{
\centering
\includegraphics[width=0.225\textwidth]{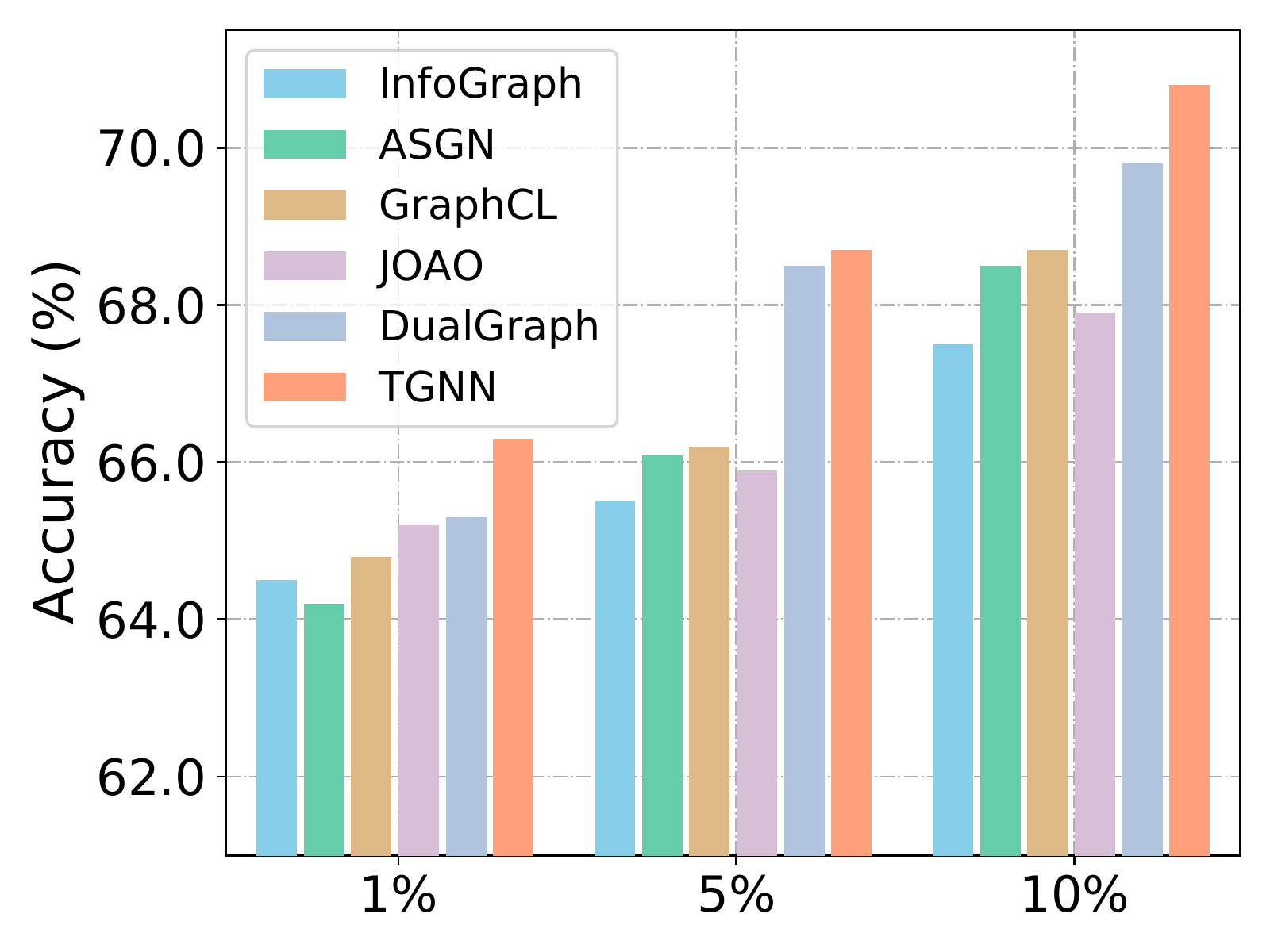}
}
\caption{Results of \method{} and baselines with different labeling ratios on two datasets PROTEINS and DD.
}
\label{fig::label_ratio}
\end{figure} 

\smallskip\noindent\textbf{Influence of Labeling Ratio.} We vary the labeling ratio of training data on PROTEINS and DD in Figure \ref{fig::label_ratio}. From the results, we can clearly see that the performance of all models generally improves as the number of available labeled data increases, demonstrating that adding more labeled data is an efficient way to boost the performance. Among these methods, \method{} achieves the best results, showing that sufficiently integrating graph topology information from complementary views can indeed benefit the performance.



\begin{table}[t]
\renewcommand\arraystretch{0.9}
\centering
\tabcolsep=3.5pt
\begin{tabular}{lccc}
\toprule 
{Methods}   & {PROTEINS} & {IMDB-B} & {REDDIT-B} \\
\midrule 
MP-Sup  & $ 63.3\pm1.4 $  & $ 63.4\pm2.1 $ & $ 69.8\pm1.1 $ \\
GK-Sup  & $ 62.6\pm0.8 $  & $ 55.4\pm1.7 $  & $ 65.3\pm0.6 $ \\
MP-Ensemble  & $ 68.1\pm1.5 $  & $ 69.8\pm1.2 $ & $ 73.7\pm1.1 $ \\
GK-Ensemble  & $ 66.9\pm1.7 $  & $ 60.5\pm1.8 $  & $ 70.8\pm0.8 $ \\
\method{} w/o Aug  & $ 69.3\pm0.8 $  & $ 71.7\pm1.3 $ & $ 75.8\pm0.7 $  \\
\midrule 
\specialrule{0em}{1pt}{1pt}
\method{} (Ours)  &  \textbf{71.0 $\pm$ 0.7}   &  \textbf{72.8 $\pm$ 1.7}  &  \textbf{76.3 $\pm$ 1.3}  \\
\bottomrule
\end{tabular}
\caption{Ablation study of \method{} with its variants.}
\label{tab::ablation}
\end{table}

\subsection{Ablation Study}
We investigate a few variants to demonstrate the effect of every part of our model: 
(1) \textbf{MP-Sup} trains an MPNN (i.e., $f_{\theta}$) solely on labeled in a supervised manner. 
(2) \textbf{GK-Sup} trains a graph kernel neural network (i.e., $g_{\phi}$) solely on labeled data. 
(3) \textbf{MP-Ensemble} replaces the graph kernel module with another message passing module with different initialization.
(4) \textbf{GK-Ensemble} replaces the message passing module with another graph kernel module with different initialization. 
(5) \textbf{\method{} w/o Aug.} removes the graph augmentation strategy before feeding graphs into the twin modules.

We present the results of different model variants in Table~\ref{tab::ablation}. First, we can clearly observe that on most datasets, MP-Sup outperforms GK-Sup. Maybe the reason is that the message passing module can utilize node attributes while the graph kernel module fails. Second, MP-Ensemble (GK-Ensemble) outperforms MP-Sup (GK-Sup), indicating our consistency learning framework can improve the model through ensemble learning. Third, our full model beats both two ensemble models, indicating the superiority of exploring similarity from complementary views.
Finally, with the graph augmentation, our full model achieves better performance, showing that graph augmentation can produce generalized graph representations beneficial to graph classification.

\subsection{Hyper-parameter Study}
\begin{figure}[t]\small
\centering
\subfigure{
\centering
\includegraphics[width=0.22\textwidth]{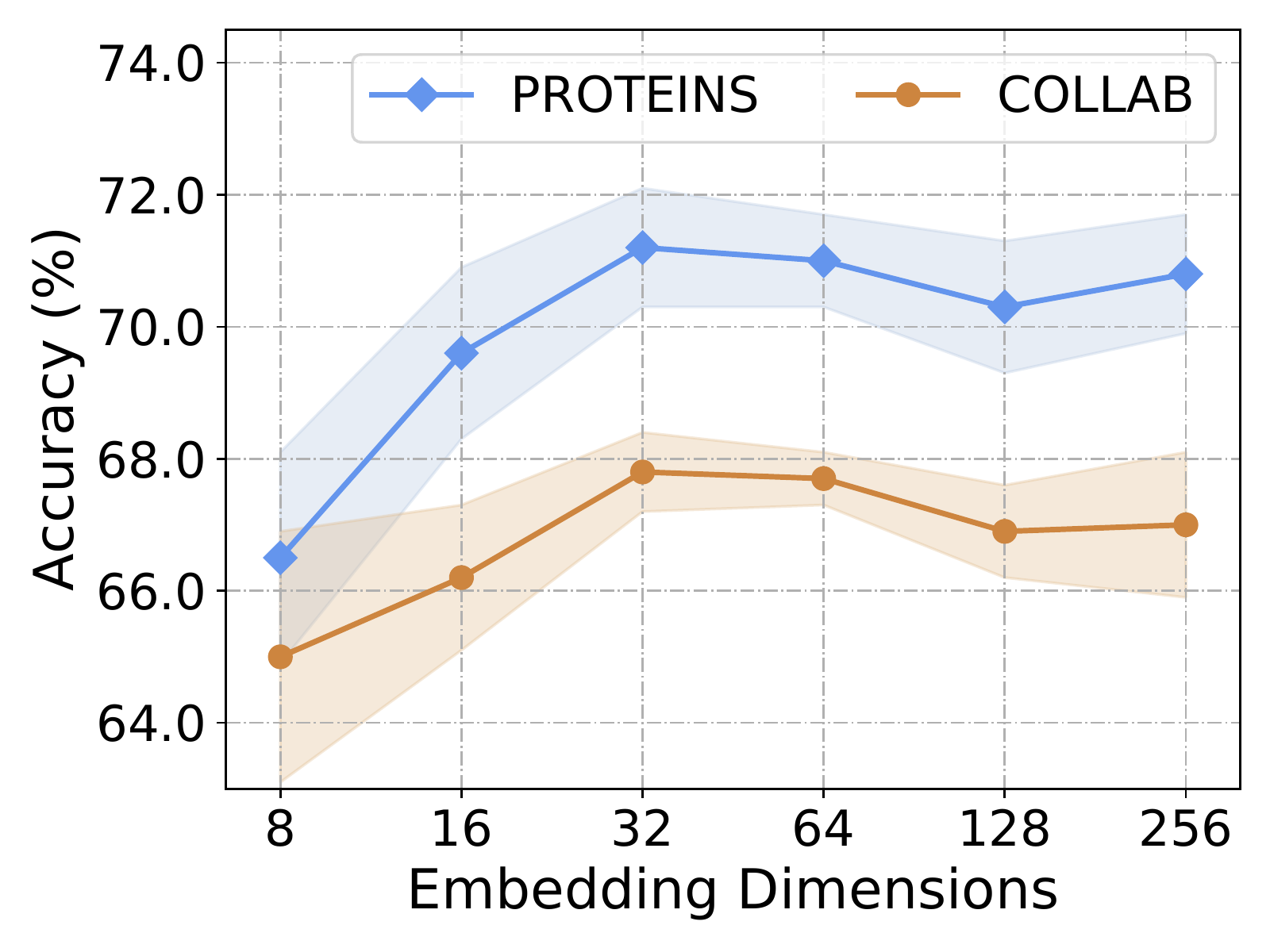}
}
\hspace{-2mm}
\subfigure{
\centering
\includegraphics[width=0.22\textwidth]{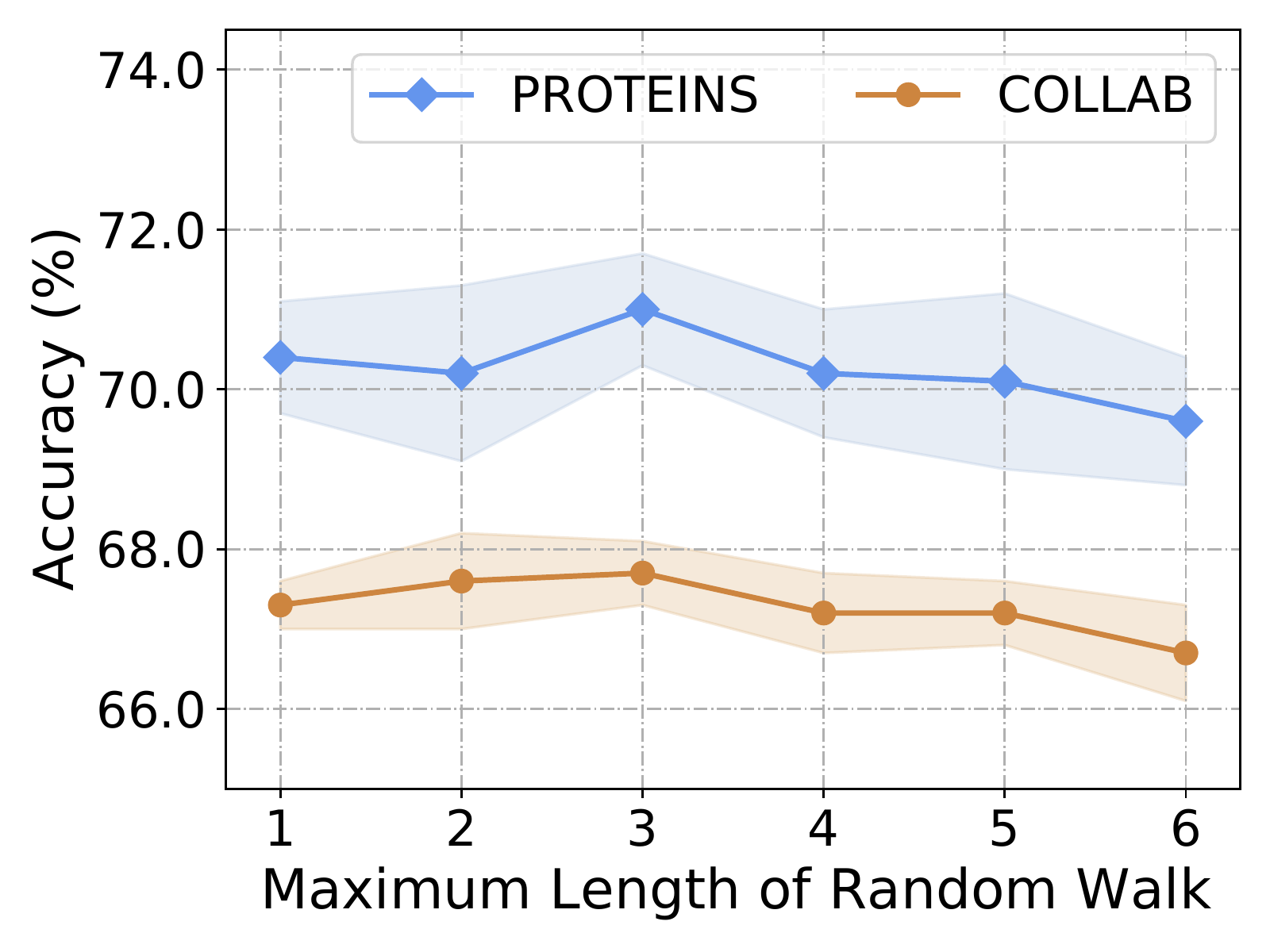}
}
\caption{Hyper-parameter sensitivity study of \method{}.}
\label{fig::parameter}
\end{figure}

We further examine the hyper-parameter sensitivity of our \method{} w.r.t. different embedding dimensions of hidden layers $d$ and the maximum length of random walk $P$ as shown in Figure \ref{fig::parameter}.
We first vary $d$ in $\{8, 16, 32, 64, 128\}$ with all other hyper-parameters fixed. It is clear that the performance saturates as the embedding dimensions reach around $64$. This is because a larger dimensionality brings a stronger representation ability at the early stage, but might lead to overfitting as the continuously increasing of $d$.
We further vary $P$ in $\{1, 2, 3, 4, 5, 6\}$ while fixing all other hyper-parameters. We observe that the performance tends to first increase and then decrease. A too-small $P$ would lead to limited topological exploration space while a large $P$ may introduce instability and fail to distinguish graph similarity.


\subsection{Case Study}
\begin{figure}[t]
	\centering
	\includegraphics[width=1.0\linewidth]{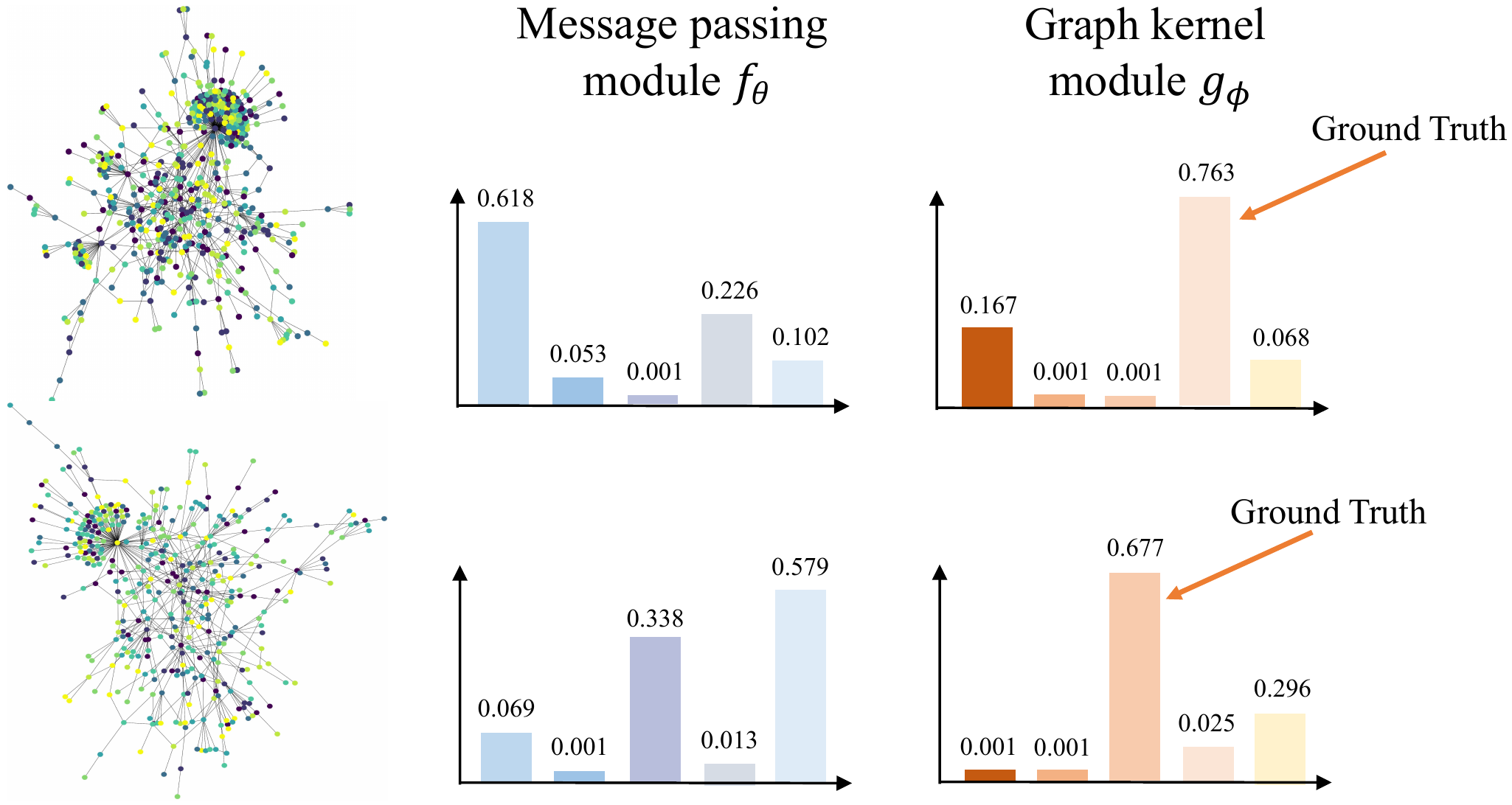}
	\caption{Visualize the power of graph kernel on REDDIT-M-5k.}
	\label{fig::case_study}
\end{figure}

\begin{figure}[t]
	\centering
	\includegraphics[width=1\linewidth]{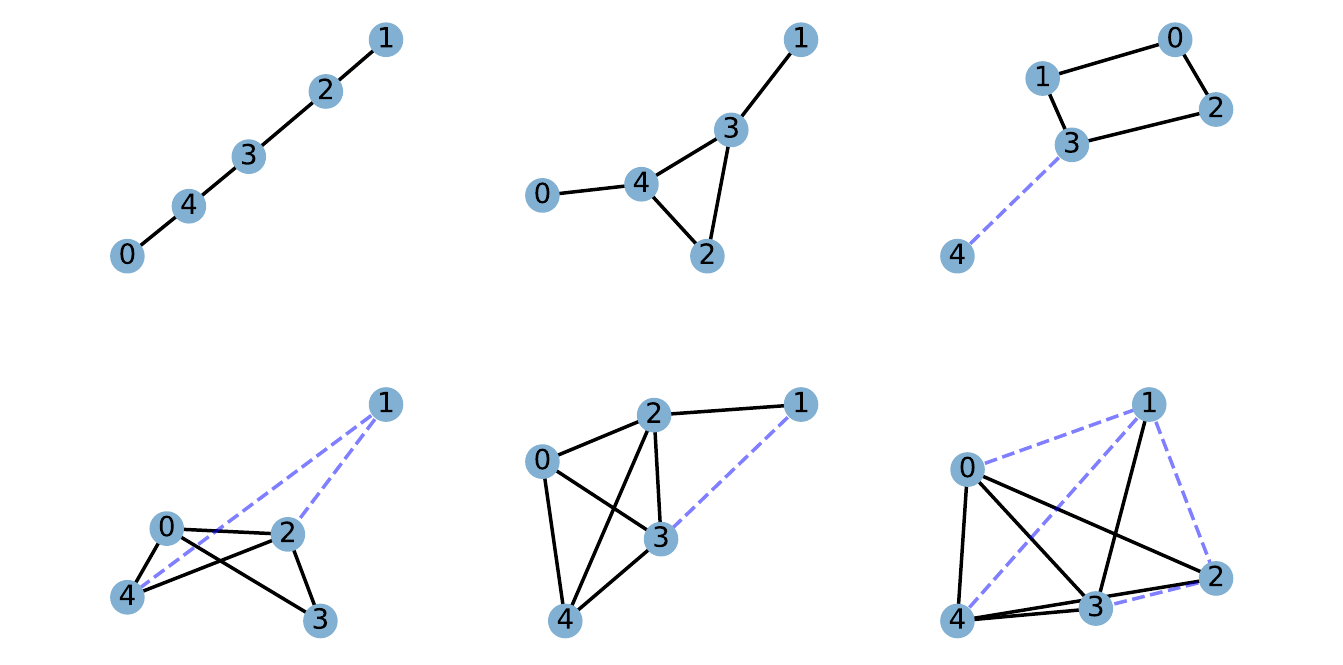}
	\caption{Illustration of hidden graphs extracted from our graph kernel module on PROTEINS.}
	\label{fig::hidden_graph}
\end{figure}

We investigate the power of our graph kernel module to show the superiority of explicitly exploring topology information. 
First, we show two cases on REDDIT-M-5k in Figure \ref{fig::case_study}, where two unlabeled examples are annotated with both the message passing module and graph kernel module. We can find that both samples are classified into wrong categories by $f_\theta$, while $g_\phi$ revises the wrong prediction successfully, which demonstrates that our graph kernel module is an important supplement for graph classification. The potential reason is that the graph kernel module is able to capture clearer topology information embedded in graph substructures. 
Furthermore, we visualize several hidden graphs derived from the graph kernel module on PROTEINS. From Figure \ref{fig::hidden_graph}, we can clearly see that our module is able to generate various kinds of graph substructures, so as to effectively capture the rich topological information in the dataset.

\section{Conclusion}
\label{sec::conclusion}
This paper presents a Twin Graph Neural Network (TGNN), a joint optimization framework that combines the advantages of graph neural networks and graph kernels. \method{} consists of twin modules named message passing module and graph kernel module, which explore graph topology information from complementary views. Furthermore, a semi-supervised framework is proposed to encourage the consistency between two similarity distributions in different embedding spaces. Extensive empirical studies show that our approach outperforms competitive baselines by a large margin.

\section*{Acknowledgments}
This paper is partially supported by National Key Research and Development Program of China with Grant No. 2018AAA0101902 as well as the National Natural Science Foundation of China (NSFC Grant No. 62106008 and No. 62006004). 

\clearpage
\balance
\bibliographystyle{named}
\bibliography{ijcai22}

\begin{thebibliography}{}

\bibitem[\protect\citeauthoryear{Adhikari \bgroup \em et al.\egroup
  }{2018}]{adhikari2018sub2vec}
Bijaya Adhikari, Yao Zhang, Naren Ramakrishnan, and B~Aditya Prakash.
\newblock Sub2vec: Feature learning for subgraphs.
\newblock In {\em PAKDD}, 2018.

\bibitem[\protect\citeauthoryear{Chen \bgroup \em et al.\egroup
  }{2020}]{chen2020convolutional}
Dexiong Chen, Laurent Jacob, and Julien Mairal.
\newblock Convolutional kernel networks for graph-structured data.
\newblock In {\em ICML}, 2020.

\bibitem[\protect\citeauthoryear{Gilmer \bgroup \em et al.\egroup
  }{2017}]{gilmer2017neural}
Justin Gilmer, Samuel~S Schoenholz, Patrick~F Riley, Oriol Vinyals, and
  George~E Dahl.
\newblock Neural message passing for quantum chemistry.
\newblock In {\em ICML}, 2017.

\bibitem[\protect\citeauthoryear{Grandvalet and
  Bengio}{2005}]{grandvalet2005semi}
Yves Grandvalet and Yoshua Bengio.
\newblock Semi-supervised learning by entropy minimization.
\newblock In {\em NeurIPS}, 2005.

\bibitem[\protect\citeauthoryear{Hao \bgroup \em et al.\egroup
  }{2020}]{hao2020asgn}
Zhongkai Hao, Chengqiang Lu, Zhenya Huang, Hao Wang, Zheyuan Hu, Qi~Liu, Enhong
  Chen, and Cheekong Lee.
\newblock Asgn: An active semi-supervised graph neural network for molecular
  property prediction.
\newblock In {\em KDD}, 2020.

\bibitem[\protect\citeauthoryear{Hinton \bgroup \em et al.\egroup
  }{2014}]{hinton2015distilling}
Geoffrey Hinton, Oriol Vinyals, and Jeff Dean.
\newblock Distilling the knowledge in a neural network.
\newblock In {\em NeurIPS Workshop}, 2014.

\bibitem[\protect\citeauthoryear{Ju \bgroup \em et al.\egroup
  }{2022}]{ju2022ghnn}
Wei Ju, Xiao Luo, Zeyu Ma, Junwei Yang, Minghua Deng, and Ming Zhang.
\newblock Ghnn: Graph harmonic neural networks for semi-supervised graph-level
  classification.
\newblock {\em Neural Networks}, 2022.

\bibitem[\protect\citeauthoryear{Kashima \bgroup \em et al.\egroup
  }{2003}]{kashima2003marginalized}
Hisashi Kashima, Koji Tsuda, and Akihiro Inokuchi.
\newblock Marginalized kernels between labeled graphs.
\newblock In {\em ICML}, 2003.

\bibitem[\protect\citeauthoryear{Kipf and Welling}{2017}]{kipf2017semi}
Thomas~N Kipf and Max Welling.
\newblock Semi-supervised classification with graph convolutional networks.
\newblock In {\em ICLR}. 2017.

\bibitem[\protect\citeauthoryear{Kojima \bgroup \em et al.\egroup
  }{2020}]{kojima2020kgcn}
Ryosuke Kojima, Shoichi Ishida, Masateru Ohta, Hiroaki Iwata, Teruki Honma, and
  Yasushi Okuno.
\newblock kgcn: a graph-based deep learning framework for chemical structures.
\newblock {\em Journal of Cheminformatics}, 2020.

\bibitem[\protect\citeauthoryear{Laine and Aila}{2017}]{laine2017temporal}
Samuli Laine and Timo Aila.
\newblock Temporal ensembling for semi-supervised learning.
\newblock In {\em ICLR}, 2017.

\bibitem[\protect\citeauthoryear{Lee and others}{2013}]{lee2013pseudo}
Dong-Hyun Lee et~al.
\newblock Pseudo-label: The simple and efficient semi-supervised learning
  method for deep neural networks.
\newblock In {\em ICML Workshop}, 2013.

\bibitem[\protect\citeauthoryear{Li \bgroup \em et al.\egroup
  }{2019}]{li2019semi}
Jia Li, Yu~Rong, Hong Cheng, Helen Meng, Wenbing Huang, and Junzhou Huang.
\newblock Semi-supervised graph classification: A hierarchical graph
  perspective.
\newblock In {\em WWW}, 2019.

\bibitem[\protect\citeauthoryear{Long \bgroup \em et al.\egroup
  }{2021}]{long2021theoretically}
Qingqing Long, Yilun Jin, Yi~Wu, and Guojie Song.
\newblock Theoretically improving graph neural networks via anonymous walk
  graph kernels.
\newblock In {\em WWW}, 2021.

\bibitem[\protect\citeauthoryear{Lu \bgroup \em et al.\egroup
  }{2019}]{lu2019molecular}
Chengqiang Lu, Qi~Liu, Chao Wang, Zhenya Huang, Peize Lin, and Lixin He.
\newblock Molecular property prediction: A multilevel quantum interactions
  modeling perspective.
\newblock In {\em AAAI}, 2019.

\bibitem[\protect\citeauthoryear{Luo \bgroup \em et al.\egroup
  }{2022}]{luo2022dualgraph}
Xiao Luo, Wei Ju, Meng Qu, Chong Chen, Minghua Deng, Xian-Sheng Hua, and Ming
  Zhang.
\newblock Dualgraph: Improving semi-supervised graph classification via dual
  contrastive learning.
\newblock In {\em ICDE}, 2022.

\bibitem[\protect\citeauthoryear{Miyato \bgroup \em et al.\egroup
  }{2018}]{miyato2018virtual}
Takeru Miyato, Shin-ichi Maeda, Masanori Koyama, and Shin Ishii.
\newblock Virtual adversarial training: a regularization method for supervised
  and semi-supervised learning.
\newblock {\em IEEE TPAMI}, 2018.

\bibitem[\protect\citeauthoryear{Narayanan \bgroup \em et al.\egroup
  }{2017}]{narayanan2017graph2vec}
Annamalai Narayanan, Mahinthan Chandramohan, Rajasekar Venkatesan, Lihui Chen,
  Yang Liu, and Shantanu Jaiswal.
\newblock graph2vec: Learning distributed representations of graphs.
\newblock {\em arXiv preprint arXiv:1707.05005}, 2017.

\bibitem[\protect\citeauthoryear{Nikolentzos and
  Vazirgiannis}{2020}]{nikolentzos2020random}
Giannis Nikolentzos and Michalis Vazirgiannis.
\newblock Random walk graph neural networks.
\newblock In {\em NeurIPS}, 2020.

\bibitem[\protect\citeauthoryear{Shervashidze \bgroup \em et al.\egroup
  }{2011}]{shervashidze2011weisfeiler}
Nino Shervashidze, Pascal Schweitzer, Erik~Jan Van~Leeuwen, Kurt Mehlhorn, and
  Karsten~M Borgwardt.
\newblock Weisfeiler-lehman graph kernels.
\newblock {\em JMLR}, 2011.

\bibitem[\protect\citeauthoryear{Spielman}{2007}]{spielman2007spectral}
Daniel~A Spielman.
\newblock Spectral graph theory and its applications.
\newblock In {\em FOCS}, 2007.

\bibitem[\protect\citeauthoryear{Sun \bgroup \em et al.\egroup
  }{2020}]{sun2020infograph}
Fan-Yun Sun, Jordan Hoffmann, Vikas Verma, and Jian Tang.
\newblock Infograph: Unsupervised and semi-supervised graph-level
  representation learning via mutual information maximization.
\newblock In {\em ICLR}, 2020.

\bibitem[\protect\citeauthoryear{Tarvainen and
  Valpola}{2017}]{tarvainen2017mean}
Antti Tarvainen and Harri Valpola.
\newblock Mean teachers are better role models: Weight-averaged consistency
  targets improve semi-supervised deep learning results.
\newblock In {\em NeurIPS}, 2017.

\bibitem[\protect\citeauthoryear{Veli{\v{c}}kovi{\'c} \bgroup \em et al.\egroup
  }{2017}]{velivckovic2018graph}
Petar Veli{\v{c}}kovi{\'c}, Guillem Cucurull, Arantxa Casanova, Adriana Romero,
  Pietro Lio, and Yoshua Bengio.
\newblock Graph attention networks.
\newblock In {\em ICLR}. 2017.

\bibitem[\protect\citeauthoryear{Vishwanathan \bgroup \em et al.\egroup
  }{2010}]{vishwanathan2010graph}
S~Vichy~N Vishwanathan, Nicol~N Schraudolph, Risi Kondor, and Karsten~M
  Borgwardt.
\newblock Graph kernels.
\newblock {\em JMLR}, 2010.

\bibitem[\protect\citeauthoryear{Xu \bgroup \em et al.\egroup
  }{2019}]{xu2019powerful}
Keyulu Xu, Weihua Hu, Jure Leskovec, and Stefanie Jegelka.
\newblock How powerful are graph neural networks?
\newblock In {\em ICLR}, 2019.

\bibitem[\protect\citeauthoryear{Yanardag and
  Vishwanathan}{2015}]{yanardag2015deep}
Pinar Yanardag and SVN Vishwanathan.
\newblock Deep graph kernels.
\newblock In {\em KDD}, 2015.

\bibitem[\protect\citeauthoryear{Ying \bgroup \em et al.\egroup
  }{2018}]{ying2018hierarchical}
Zhitao Ying, Jiaxuan You, Christopher Morris, Xiang Ren, Will Hamilton, and
  Jure Leskovec.
\newblock Hierarchical graph representation learning with differentiable
  pooling.
\newblock In {\em NeurIPS}, 2018.

\bibitem[\protect\citeauthoryear{You \bgroup \em et al.\egroup
  }{2020}]{you2020graph}
Yuning You, Tianlong Chen, Yongduo Sui, Ting Chen, Zhangyang Wang, and Yang
  Shen.
\newblock Graph contrastive learning with augmentations.
\newblock In {\em NeurIPS}, 2020.

\bibitem[\protect\citeauthoryear{You \bgroup \em et al.\egroup
  }{2021}]{you2021graph}
Yuning You, Tianlong Chen, Yang Shen, and Zhangyang Wang.
\newblock Graph contrastive learning automated.
\newblock In {\em ICML}, 2021.

\end{thebibliography}

\end{document}